\documentclass{llncs}
\usepackage{latexsym}
\usepackage{color, soul}
\usepackage{amssymb}
\usepackage{graphicx}
\usepackage{amsmath}
\usepackage{array}
\usepackage{tabu}
\usepackage{multirow}
\usepackage{multicol}
\usepackage{caption}
\usepackage{subfig}
\begin{document}

\title{Aspect Specific Opinion Expression Extraction using Attention based LSTM-CRF Network}
\titlerunning{Aspect Specific Opinion Expression Extraction using Attention based LSTM-CRF Network}  
%
\author{Abhishek Laddha\inst{1} \and Arjun Mukherjee\inst{2}}
\authorrunning{Laddha and Mukherjee} 
\institute{Indian Institute of Technology Delhi, India - 110016\\
\email{laddhaabhishek11@gmail.com}
\and
Department of Computer Science, University of Houston, TX, USA\\
\email{arjun@cs.uh.edu}}

\maketitle              
\begin{abstract}
Opinion phrase extraction is one of the key tasks in fine-grained sentiment analysis. While opinion expressions could be generic subjective expressions, aspect specific opinion expressions contain both the aspect as well as the opinion expression within the original sentence context. In this work, we formulate the task as an instance of token-level sequence labeling. When multiple aspects are present in a sentence, detection of opinion phrase boundary becomes difficult and label of each word depend not only upon the surrounding words but also with the concerned aspect. We propose a neural network architecture with bidirectional LSTM (Bi-LSTM) and a novel attention mechanism. Bi-LSTM layer learns the various sequential pattern among the words without requiring any hand-crafted features. The attention mechanism captures the importance of context words on a particular aspect opinion expression when multiple aspects are present in a sentence via location and content based memory. A Conditional Random Field (CRF) model is incorporated in the final layer to explicitly model the dependencies among the output labels. Experimental results on Hotel dataset from Tripadvisor.com showed that our approach outperformed several state-of-the-art baselines.
\end{abstract}

\section{Introduction} 
Aspect based sentiment analysis \cite{liu2012sentiment} is one of the main frameworks for fine-grained sentiment analysis and is used in several downstream tasks such as opinion summarization, extracting opinion targets, opinion holders, opinion expressions etc. One of the main goals of aspect based sentiment analysis is to identify the fine-grained product properties (aspects) and their opinion. In \cite{titov2008,jo2011,liu2015fine} the aspect term and opinion words are jointly extracted but lack correspondence between the aspect and opinion terms. For example, in the sentence  \textit{``the {\color{blue}\underline{food} was excellent and plentiful} and the {\color{red} \underline{waitstaff} was extremely friendly and helpful}"}, discovering aspect words as \textit{\{food, waitstaff\}} and opinion words as \textit{\{excellent, plentiful etc.\}} is definitely useful but being able to extract phrases that retain the sentence context as aspect specific opinion expressions such as (\textit{\underline{food} was excellent and plentiful}, \textit{\underline{waitstaff} was extremely friendly and helpful}) would be more expressive and provide more information about the aspect. These opinion phrases can be further used in downstream application such as aspect sentiment classification, aspect summarization.

Traditionally, subjective expression extraction \cite{choi2005,breck2007} has been formulated as a token-level sequence labeling task and has employed a CRF based approach using hand-crafted features. Recent success of distributed representation of words \cite{mikolov2013distributed,pennington2014glove} provides alternate approach to learn the continuous valued dense vectors for latent features in hidden layers. \cite{irsoy2014opinion,liu2015fine} apply deep Recurrent Neural Network (RNN) to extract the opinion expression and opinion target from sentences. They have shown that a deep RNN approach outperforms traditional CRF and semi-CRF. Approaches in \cite{irsoy2014opinion,liu2015fine} learn the opinion phrase representation based on the latent features learned of context words but are incapable of explicitly providing the cues from the aspect word. \cite{tang2016aspect,wangattention,tang2015effective} proposed models to extract the sentiment of an aspect in a sentence by taking into account the aspect words. They are mainly focusing on positive or negative sentiment instead of generic opinion phrase about aspect.  

In this paper, we present a neural network architecture with Bi-LSTM and an attention mechanism to take into account the aspect cues. Bi-LSTM layer learns the various sequential patterns among the words without requiring any hand-crafted features. Most of the current work in aspect classification \cite{tang2016aspect,wangattention,tang2015effective} assumes presence of one aspect in the sentence. If there are multiple aspects in the same sentence they consider them as separate instance ignoring the effect of one aspect on another. We believe that if there are multiple aspects in a sentence, explicitly feeding the importance of context word based on the content and location for a particular aspect is an essential signal to decide whether a context word is in opinion expression of aspect or not. 

Inspired by recent success of attention based computational model as in aspect sentiment classification \cite{tang2016aspect,wangattention}, machine translation \cite{bahdanau2014neural}, we propose an attention mechanism which takes into account the multiple aspects in the sentence based on the context/surrounding word’s location from multiple aspect word. 
This layer would be helpful in tagging the words which are in between the two aspect and can be included into both aspect opinion expression thereby helping in locating the precise boundary of aspect specific opinion phrases. A CRF model is incorporated in the final layer to explicitly model the dependencies among the output labels.\\
\section{Related Work}
We briefly review the existing studies on subjective expression extraction task and aspect-based sentiment analysis using neural networks in this section
\subsection{Subjective expression extraction}
Early works on fine-grained opinion extraction \cite{choi2005,breck2007} have used various parsing, syntactic, lexical and dictionary based features to extract a subjective expression employing a CRF based approach. Various features based on dependency relations \cite{johansson2011} and opinion lexicon have been used for opinion expression extraction. Further, \cite{yang2012,yang2013} employed semi-CRFs which allowed sequence labeling at the segment level. \cite{yang2013} proposed a joint inference model to jointly detect opinion expressions, opinion holders and target as well as relation among them. While they have made important progresses, their performances mainly rely on rich hand crafted features and other pre-processing steps such as dependency parsing.\par
There have been work exploring the combinations of sequential neural network (e.g. LSTM, RNN) on sequence labeling tasks such as Named Entity Recognition (NER), language understanding. \cite{huang2015bidirectional,yao2014recurrent,lample2016neural} added a CRF layer on top of RNN network and showed performance improvement on Named Entity Recognition (NER) and language understanding. \cite{ma-hovy} extended above model by using CNN on character of words to get word level representation. These works have mostly explored the neural networks in NER as opposed to opinion phrase extraction. \par
The works in \cite{laddha2016,wang2016mining}, are the closest to ours as they focus on aspect specific opinion terms. While \cite{wang2016mining} does not discover phrases, \cite{laddha2016} employs higher order CRF features for phrase extraction and is considered as a baseline.

\subsection{Neural network for Aspect based sentiment analysis}
Recent studies have shown that deep learning models can automatically learn the inherent semantics and syntactic information from data and this achieves better performance for sentiment analysis. Regarding  aspect based sentiment analysis \cite{wang2016recursive,liu2015fine,yin2016unsupervised} models target aspect term extraction as a sequence tagging task using neural network. In \cite{liu2015fine}, RNN and word embedding were combined to extract explicit aspects. In \cite{wang2016recursive}, recursive neural network based on dependency tree and CRF were integrated in a unified framework to extract the aspect and opinion terms. \cite{yin2016unsupervised} used word and dependency paths embeddings as features in CRF. These methods are mostly focused on aspect term extraction instead of aspect specific opinion expression.\par
Also related are the works around aspect based sentiment classification \cite{tang2016aspect,wangattention,tang2015effective} and the work in \cite{nguyen2015phrasernn} which proposed an extension of RNN to identify the aspect’s sentiment. \cite{wangattention} proposed an LSTM model with attention mechanism which focuses on different part of sentences given the aspect input. Further, in \cite{tang2016aspect}, a deep memory network was proposed for explicitly capturing the importance of each context word when inferring the polarity of given aspect. These approaches provide the sentiment about the aspect but do not give in-depth information about the aspect such as the associated opinion expression. 

\section{Background: LSTM-CRF model}
This section briefly describes bidirectional LSTM-CRFs that lays the foundation for the proposed attention based LSTM-CRF network. For more details, refer to \cite{huang2015bidirectional,lample2016neural,ma-hovy}.
\subsection{Bidirectional LSTM Network}
Recurrent Neural Networks (RNNs) are a family of neural networks that take an input $X = \{x_1, \ldots, x_T \}$ and yield the sequence hidden representation $\{ h_1, \ldots, h_T \}$ where each $h_t \in \mathbb{R}^{d \times 1}$  represents the semantic information of the left context of $x_t$. In practice these models fail to capture the long term dependencies and suffer from the vanishing gradient problem. \cite{hochreiter1997long} proposed an LSTM cell which eliminates these problems by employing several gates to control the information flow in the cell. For each word of a given sentence, an LSTM computes a representation $\overrightarrow{h_t}$ of the left context of sentence. Similarly, a right context information $\overleftarrow{h_t} $ also contains the useful information. This can be achieved by employing another LSTM network which reads the current sentence in the reverse order. Now, the word representation in this bidirectional LSTM would be a concatenation of its left and right latent context representations $ h_t = [\overrightarrow{h_t}; \overleftarrow{h_t}] \in \mathbb{R}^{2d \times 1}$. 
\subsection{LSTM-CRF Model}    
For the task of sequence tagging, a very simple approach would be to predict the labels $y_t$ independently for each token using a simple feed-forward neural network classification model which takes the output of the LSTM as input vectors. But for labeling of opinion expressions, there is a strong dependency associated with the previous and current label. Therefore, instead of predicting label for each token independently, we model them jointly using CRF \cite{lafferty2001}. Let's consider $P \in \mathbb{R}^{T \times q}$ to be a matrix of the output of a Bi-LSTM after projecting it on to a linear layer whose size is equal to number of distinct labels $q$, $T$ is the number of tokens. We also define a transition matrix $A \in \mathbb{R}^{q \times q}$ where each entry $a_{i,j}$ represents a probability of transition from state $i$ to $j$ of consecutive output labels. Then, score for a complete sentence can be defined as follows 
\belowdisplayskip=3pt
\abovedisplayskip=3pt
\begin{equation}
\label{score}
s(X,y) = \sum_{t=0}^{T} A_{y_t,y_{t+1}} + \sum_{t=1}^{T} P_{t,y_t}
\end{equation}
where start state and end state is added in the sentence. Since we are considering only bigram interactions among labels, dynamic programming \cite{lafferty2001} can be used to compute $A$ and the best possible sequence labels during inference \cite{collobert2011natural}.
\section{Attention based LSTM-CRF network (LSTM-ATT-CRF)}
\subsection{Task Definition and Notation}
\begin{figure}[t!]
\includegraphics[width=0.9\textwidth, height=0.4\textheight]{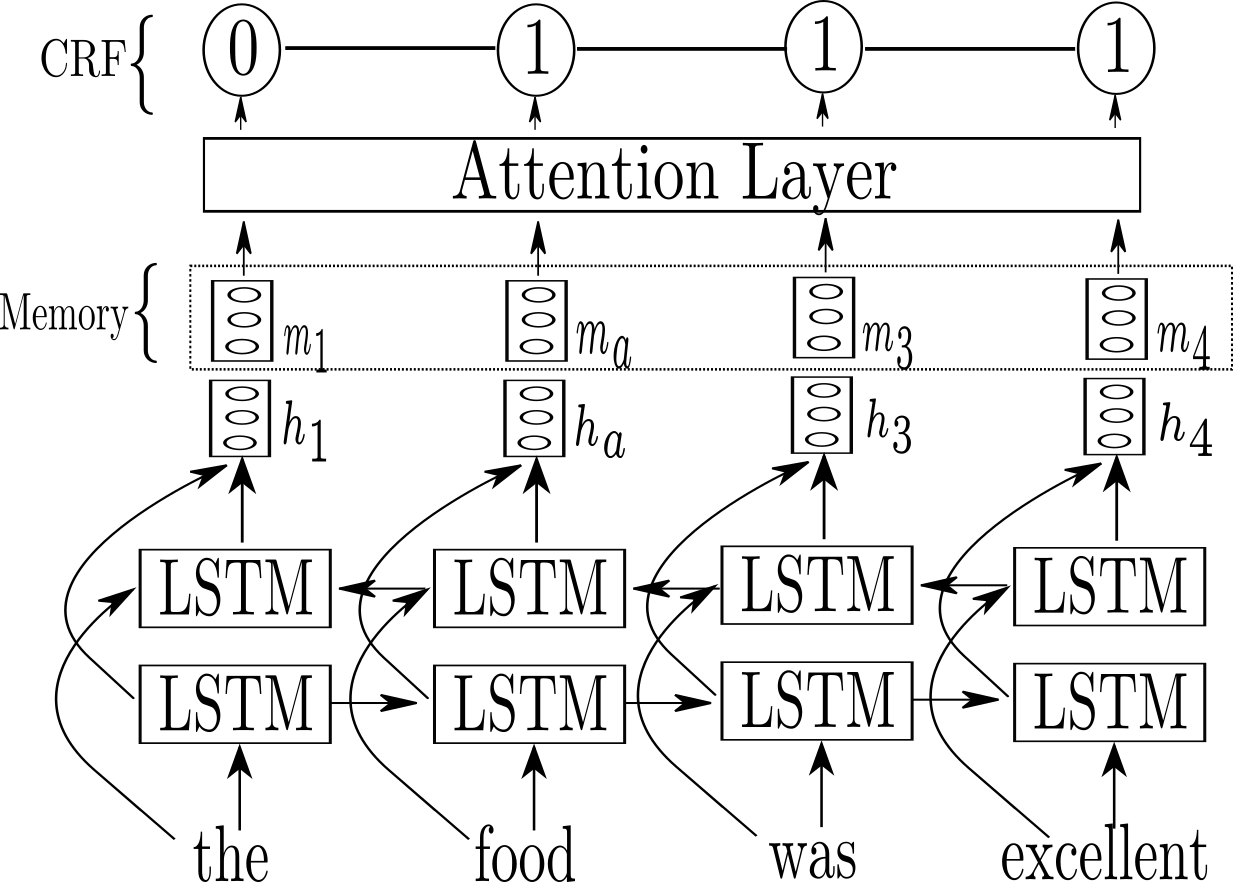}
\caption{\label{arch} Main Architecture of network. Word input are given as word embedding to Bi-LSTM. 
}
\end{figure}

Given a sentence $X=(w_{1},\ldots w_{T})$ with $T$ number of words and set of $n$ aspect words $ \{ w_{a_1} ,\ldots w_{a_n} \}$ where ${a_k} \in [1,T]$ is mentioned in sentence $X$. Our task is to extract set of relevant opinion expression containing the aspect and opinion phrase. We formulated this as sequence labeling problem where each word of a sentence has label $y_t \in \{0,1\}$. $y_t$ is assigned to 1 if a word is in any aspect specific opinion expression or 0 otherwise. To represent each word, we map it onto a low dimensional continuous vector and corresponding to each word $w_t$, it's embedding is represented as $v_t \in \mathbb{R}^{d_w \times 1}$ where $d_w$ is the word embedding size. The complete architecture of network is shown in Figure \ref{arch} where Bi-Directional LSTM is similar to network described in the previous section.  
\subsection{Attention Network}
The basic idea behind attention is to assign importance to each context word based on the latent representation and memory. In our setting, memory refers to the multi aspect information present in a sentence. Using the location of the context word from the multiple aspects, memory vector $m_t$ is computed using Eq. \ref{mem}. Main intuition is that every aspect doesn't contribute equally to determine the polarity of each context word in the sentence. The words which are distant from the particular aspect word would influence less by that aspect word. In this work, we define the location of context words as it's absolute distance with the aspect in the original sentence. Memory vector for each token at $t$ is defined 
\begin{equation}
\label{mem}
m_t = \frac{\sum_{k=1}^{n} v_{a_k} \left( 1 - \frac{l_{t,k}}{T} \right)}{n}
\end{equation}
\begin{equation}
\label{weight}
g_t = tanh \left( W_{attn} \left[ m_t; h_t \right] + b_{attn} \right)
\end{equation}
where $l_{t,k}$ is number of word between $w_t$ and $w_{a_k}$, $W_{attn} \in \mathbb{R}^{(2d + d_w) \times 1}$, $v_{a_{k}}$ is the embedding vector of aspect $a_{k}$. Based on the memory vector and hidden representation, the model assigns an score $g_t$ to each context word using Eq. \ref{weight} which takes into account the relation between word and multiple aspects. After getting the $g_t$'s we feed them to softmax function to get the importance weights $\alpha_t$ for each context word, such that $\sum_t \alpha_t = 1$ and each $\alpha_t \in [0,1]$.
\begin{table*}[t]
\begin{center}
\resizebox{\columnwidth}{!}{
\begin{tabular}{|c@{ }c@{ }c@{  }c@{  }c@{  }c@{  }c@{  }c@{  }c@{  }c@{  }c@{  }c|}
\hline
& {\color{blue}food} & {\color{blue}was} & {\color{blue}excellent} & {\color{blue}and} & {\color{blue}plentiful} & and & the & {\color{red}waitstaff} & {\color{red}was} & {\color{red}extremely} & {\color{red}friendly}\\
Before &attention: &   &  &  &  &  &  &  &  &  &  \\
$P(y = 1)$ & 0.6 & 0.43  & 0.79 & 0.64 & \textbf{0.46} & 0.3 & 0.2 & 0.6 & 0.45 & 0.76 & 0.82\\
$P(y = 0)$ & 0.4 & 0.57  & 0.21 & 0.36 & \textbf{0.54} & 0.7 & 0.8 & 0.4 & 0.55 & 0.24 & 0.18\\
After &attention: &   &  &  &  &  &  &  &  &  & \\
$P(y = 1)$ & 0.7 & 0.45  & 0.86 & 0.64 & \textbf{0.56} & 0.3 & 0.2 & 0.7 & 0.47 & 0.83 & 0.9\\
$P(y = 0)$ & 0.3 & 0.55  & 0.14 & 0.36 &\textbf{0.44} & 0.7 & 0.8 & 0.3 & 0.53 & 0.17 & 0.1\\
\hline
\end{tabular}
}
\end{center}
\captionof{figure}{\label{example} An illustration of example of our neural attention network for aspect specific opinion labeling. Words in blue and red corresponds to opinion expression about ``food'', and``waitstaff'' respectively.}
\vspace{-10pt}
\end{table*}
\begin{equation}
\label{linear}
P_t = W_{linear}^{Tr} (\alpha_t \times h_t) + b_{linear}
\end{equation} 
A linear layer is applied to transform the hidden representation vector to the scores of each output tag using the Eq. \ref{linear}, where $W_{linear} \in \mathbb{R}^{2d \times q}$ and $b_{linear} \in \mathbb{R}^{q \times 1}$ and after that score of a sequence was calculated using Eq. \ref{score}. Here, $P_{t, y_{t}}$ is unscaled probability of word $w_t$ having label $y_t$. In the absence of attention $\alpha_t $ will become $1$ however, attention will provide weighted hidden state with respect to aspect words. Figure \ref{example} shows a example where word ``plentiful" have low confidence in inclusion of opinion expression due to long distance from aspect word ``food" and closeness to aspect word ``waitstaff" for which it doesn't express opinion. While attention will learn to give more importance to those words because there is direct interaction of hidden vector with aspect word and there are lots of opinion expression about aspect ``food''  which includes the words of similar meaning as of ``plentiful". 
\subsection{Model Training}
The model can be trained end-to end using backpropogation, where the objective function is to maximize the log-probability of correct sequence score as defined in Eq. \ref{loss}.
\begin{equation}
\label{loss}
p(y|X) = \frac{exp(s(X,y))}{\sum_{\hat{y}} exp(s(X,\hat{y}))}
\end{equation}  
where $X$ denotes the sequence of words and $y$ is the corresponding sequence label. $s(X,y)$ is score defined in Eq. \ref{score}, $P_t$ learns the probability of each label independently from Bi-LSTM while $A$ learns the dependency among the various labels. For \textit{e.g.,} in Figure \ref{example} some stopwords such as ``was'' could have low probability of label $y_t = 1$, but decoding complete sequence using Eq. \ref{score} will consider surrounding label and hence inclusion of such words in opinion expression. 
Model parameters are the LSTM weights, $W_{attn}, W_{linear}, b_{attn}, b_{linear}, A$ and word embeddings. Except word embeddings, other parameters are initialized using sampling from uniform distribution $U(-\epsilon, \epsilon)$, where $\epsilon = 0.01$.\\
\textbf{Word Embeddings}:  Word embeddings are initialized using the pre-trained embeddings. We  use  Stanford's  publicly  available  GloVe \footnote{http://nlp.stanford.edu/projects/glove/}  100-dimensional  embeddings \cite{pennington2014glove}. We also experimented with two other embeddings, namely Senna \footnote{http://ronan.collobert.com/senna/} 50-dimensional  embeddings \cite{collobert2011natural} and Google's word2vec \footnote{https://code.google.com/archive/p/word2vec/} 300-dimensional \cite{mikolov2013distributed}. Parameter $d_w$ depends on dimension of pre-trained word embedding vectors. \\
\textbf{Features}: Although NNs learn the word features (i.e. embedding) automatically, \cite{liu2015fine} shows that incorporating other linguistic features like part of speech (POS) and syntactic information (e.g., phrase chunks) helps in the training to learn a better model. We used the syntactic features (POS tags) and phrase chunk features as input in the LSTM network. Similar to word embedding, each feature is also mapped to feature embedding which gets learned during training. Input to LSTM network is a concatenation of word embedding and feature embeddings.    
\section{Experiments} 
\subsection{Dataset}
To demonstrate the effectiveness of our model we performed experiments on the Hotel dataset from Tripadvisor.com used in \cite{wang2010}. Labeling opinion phrases for each aspect is a tedious task and require lots of human effort. Further, training deep learning model generally needed substantial amount of training data. To overcome this bottleneck, we tried to label the phrases using heuristic rules. Seed words for each aspect used in \cite{wang2010} were used to label the location of aspect words in review sentences. Once, aspect words in sentence were determined, opinion expressions around those aspect words are labeled as described next.
\subsubsection*{Labeling using heuristic rules}
Since, we mainly focus on opinion expressions surrounding the aspect word, heuristic rules could be written with the help of Part of Speech (POS) tags and polarity of the surrounding words. We used the opinion lexicon\footnote{http://www.cs.uic.edu/$\sim$liub/FBS/opinion-lexicon-English.rar} derived from \cite{hu2004} for positive and negative words. Labeling of the phrase boundary surrounding the aspect word for positive opinions phrases was done as follows:\par
In the first step, we searched for the positive terms (using the sentiment lexicon) in the window of $[-5,5]$ around the aspect word. To have a compact opinion phrase, it should not include the opinion about the other aspects. So, only considering the extremes with a sequence index of $[$aspect, positive-term$]$ would not yield good phrases. Therefore, we divided the presence of positive words in two cases. First, when the positive word was before the aspect word, to capture all the opinion words in a phrase talking about noun aspect, we took the farthest adjective from the aspect word. If aspect word was verb, then we took the nearest adverb since adverbs are mostly situated immediately before the verb. Second, when the positive word is after the aspect word, we took the nearest adjective (from the aspect word) because adjectives are generally immediately followed by nouns. We also included the negative phrases with negator words in surrounding window by finding negative terms in window of $[-5,5]$ and then finding negator terms such as "not", "don't" and following the same procedure described above. This process generally yielded us the minimum boundaries of phrases but could have omitted some opinion words which we included using the method described below. \par
Next, we extended the phrase boundaries using the basic definition of adjective and adverb: (i) If first word of a phrase is coming before the aspect word and it is a verb, then we look at the word before the verb, if the word was an adverb, then we include it in that phrase, (ii) If the last word of the phrase was an adjective and the next word after that was noun, then we included all the consecutive nouns after that, (iii) If last word of the phrase was an adverb, then we included the next word if it was a verb, (iv) If last word of the phrase was noun, then we included all the consecutive nouns after that. Similar process was applied for extracting negative opinion expressions as well. We explain our procedure using the following \textit{e.g.,} ``The room provided a nice view of the lagoon", aspect word is ``room'' which is noun and opinion word is ``nice'' which is adjective. Since adjective word is after the aspect word we took the nearest adjective and extracted phrase would be ``room provided a nice'' which is not complete. Now, we extend this using the rule (ii) which will include the noun word ``view'' since it immediately follows the adjective. Thus this completes the opinion expression. We observed that following these heuristic rule are able to capture various fluid opinion expression like ``wonderful hotel at a reasonable price,'' and ``rooms do feel quite bland 
''.
\subsubsection*{Dataset Dissection} 
Using the above procedure, we labeled a total of $10,775$ sentences which was split in $80:20$ ratio for training and validation. We want to evaluate on the dataset which is completely realistic and wanted to test the ability of our model to retrieve phrases which might not have labeled using the heuristic rules. Hence, for testing, we manually labeled another disjoint set of $1,683$ sentences after getting the location of aspect labeled from the seed words. Dataset will be released to serve as language resource. We preprocessed the data by lowercasing all the word and replaced all cardinal numbers with ``NUM" symbol and removed words appearing only once. \par
\subsection{Parameters Settings}
Our model was implemented in tensorflow\footnote{http://tensorflow.org/} using the Adam optimizer with initial learning rate of $.01$ and early stopping criteria \cite{graves2013speech} was used based on validation set accuracy. The decaying learning rate for Adam was set to $0.05$. Care was taken to reduce overfitting by adding a dropout layer regularizer \cite{srivastava2014dropout} with keep probability of $0.5$ and gradients were clipped at $5$. Other hyperparamters such as dimension of the hidden states of LSTM were kept same for all model $d = 100$, $\#$ of layers as $3$, batch size as $10$, and maximum length of sentence was set to $50$ which were determined using pilot experiments. 
\subsection{Comparison with Baselines}
\begin{table}[t]
\noindent
\begin{center}
\resizebox{0.6\columnwidth}{!}{
\begin{tabular}{|c|c|c|c|}
\hline 
Model & Precision & Recall & F-score \\ \hline
CRF & 82.77 & 69.01	& 75.26	\\ \hline
semi-CRF & 84.63 & \textbf{78.27} & 81.29 \\ \hline
LSTM-CNN-CRF & 88.46	& 72.47	& 79.67 \\ \hline
LSTM-ATTN-CRF & \textbf{88.80} & 75.86 & \textbf{81.82} \\ \hline
\end{tabular}
}
\end{center}
\caption{\label{best-models} Comparison of results with baselines}
\vspace{-10pt}
\end{table}
We compared our model with the following most relevant baselines.\\
\textbf{CRF}: We use linear chain-CRF \cite{lafferty2001} and higher order features as described in \cite{laddha2016}. \\
\textbf{semi-CRF}: This is due to model in \cite{yang2012} that used dependency tree features with semi-CRF to label the sequence at segment level.\\
\textbf{LSTM-CNN-CRF}: This is due to model in \cite{ma-hovy} which used the word and character level representation using combination of LSTM, CNN and CRF for sequence labeling task. \\
\textbf{LSTM-ATT-CRF}: Our complete proposed model which have attention over the output of Bi-LSTM using aspect memory with CRF layer.\par
Next, we also explored two simplified versions of our model \\
\textbf{LSTM-ATT}: This model employs the cross-entropy between the predicted and target labels for loss instead of maximizing the CRF score.\\
\textbf{LSTM-CRF}: The concatenated output vectors of Bi-LSTM are passed directly into the linear layer for computing the CRF score.  
\subsection{Discussion}
We used word based micro precision, recall and F-score to evaluate the quality of the model. \cite{yang2013} has shown that it is difficult to get exact opinion expression boundaries even for human annotators and hence focused on precision, recall measures at word level instead of complete expression level. Precision is defined as $\frac{|C \cap P|}{P}$ and recall as $\frac{|C \cap P|}{C}$, where $C$ and $P$ are the correct and predicted set of word labels respectively.

Table \ref{best-models} illustrates the comparison results of baselines with our best model LSTM-ATT-CRF. Our model significantly outperforms CRF and LSTM-CNN-CRF on F-score.  It also improves over semi-CRF at $p<0.05$. semi-CRF performs close to our model due to the fact that many opinion phrases are noun phrases (NPs) and verb phrases (VPs), and its use of segment level labeling greatly improved recall but it suffers in precision.
\begin{table*}[t]
\small
\noindent
\begin{center}
\begin{tabular}{|c|c|c|c|c|c|c|c|c|c|}
\hline 
\multirow{2}{*}{ Model } & \multicolumn{3}{|c}{Senna} & 
\multicolumn{3}{|c}{word2vec} & \multicolumn{3}{|c|}{Glove} \\ \cline{2-10}
 & P & R & F & P & R & F & P & R & F  \\ \hline
LSTM-CNN-CRF & 88.46	& 72.47	& 79.67	& 89.93	& 71.18	& 79.46	& 87.35	& 73.15	& 79.62	\\ \hline
LSTM-ATT-CRF & \textbf{88.80} & 75.86 & \textbf{81.82}	& 88.40	& 75.08	& 81.20	& 87.73	& 76.30	& 81.62	 \\ \hline
LSTM-ATT & 87.3	& 73.31	& 79.58	& 88.22	& 74.57	& 80.82	& 87.4	& 74.84	& 80.67	 \\ \hline
LSTM-CRF & 88.14	& 75.94	& 81.59	& 88.68	& 74.13	& 80.75	& 87.98	& 75.40	& 81.2	 \\ \hline
\end{tabular}
\end{center}
\caption{\label{word-embed} Performance on aspect specific opinion expression task on Precision, Recall, F1 for different models when initialized using various pre-trained embeddings}
\vspace{-20pt}
\end{table*}
\begin{table*}[t]
\small 
\begin{center}
\resizebox{\columnwidth}{!}{
\begin{tabular}{|c@{ }c@{ }c@{  }c@{  }c@{  }c@{  }c@{  }c@{  }c@{  }c@{  }c@{  }c@{  }c@{  }c@{  }c@{  }c|}
\hline
hotel&in& [[ excellent  & \underline{location} & close & to & everything ]] & we & were & impressed & [[ excellent & \underline{service} & at & reception & upon & arrival ]] \\
0.021 & 0.013 & {\color{blue}0.048}  & {\color{blue}0.066} & {\color{blue}0.075} & 0.026 & {\color{blue}0.075} & 0.015 & 0.015 & 0.067 & {\color{blue}0.076} & {\color{blue}0.073} & {\color{red}0.06} & {\color{blue}0.050} & 0.031 & 0.026\\
\hline
\end{tabular}
}
\resizebox{\columnwidth}{!}{
\begin{tabular}{|c@{ }c@{ }c@{ }c@{ }c@{ }c@{ }c@{ }c@{ }c@{ }c@{ }c@{ }c@{ }c@{ }c@{ }c@{ }c@{ }c|}
\hline
there & are & [[ waterfalls  & in & \underline{lobby} & area ]] & and & [[ free & easy & fast & \underline{internet} & access ]] & but & only & in & \underline{lobby} & area \\
0.027 & 0.015 & {\color{red}0.0187}  & 0.024 & {\color{blue}0.107} &{\color{blue} 0.066} & {\color{red}0.083} & {\color{blue}0.103} & {\color{blue}0.106} & {\color{blue}0.103} & {\color{blue}0.04} & {\color{red}0.026} & 0.038 & 0.019& 0.016& 0.027 &0.016 \\
\hline
\end{tabular}
}
\end{center}
\caption{\label{attention} Example of attention weight for different example. Underlined words are aspect words, weights colored in blue are probably correct while weights in red are wrong. True opinion expression are included in [[ ]]. Higher weights mean the more probable a word is in opinion expression.
}
\vspace{-10pt}
\end{table*}

Further from Table \ref{word-embed}, we note LSTM-ATTN-CRF outperforms character based LSTM-CNN-CRF and LSTM-CRF across all word embedding  which shows including the aspect information using the attention is effective even when features based on POS and syntactic tags (phrase chunk units) are included as input. Our complete model outperform LSTM-ATT significantly which shows that adding the CRF layer to capture the dependency among output label is useful. Also interesting to note is that Senna embeddings perform best for aspect specific opinion expression. This is due to the fact that Senna was trained on various NLP task such as NER, POS and SRL whereas other Glove, word2vec were trained on general co-occurence of words.

For a yet deeper understanding of the attention mechanism, Table \ref{attention} shows the attention weights for two examples. Weights of context words around aspect confirms that our attention mechanism is able to assign the weights based on the word and aspect. Reason for incorrect word such as `waterfall' is due to their low occurrences in the corpus while others are stopwords which sometimes get included in the opinion expression. Our model is able to assign the substantial weight to many neutral words such as `close' and `everything' based on the aspect which contributes to its effectiveness over other baselines.
\section{Conclusion}
In this paper, we presented an attention based LSTM-CRF (LSTM-ATT-CRF) for aspect specific opinion expression task. The model works for both single and multiple aspect sentences and improves phrase discovery by leveraging the latent interactions among the aspect and opinion words based on the content and location which we modeled via attention mechanism. Experimental results on a hotel dataset showed superior performance over several baselines. The work also produced a labeled dataset which shall be released as a resource.

\section*{Acknowledgement}
This work is supported in part by NSF 1527364.
\bibliographystyle{splncs03}

\end{document}